\pdfoutput=1

\documentclass[11pt]{article}

\usepackage{acl}

\usepackage{times}
\usepackage{latexsym}
\usepackage{booktabs}
\usepackage{amsfonts}
\usepackage{amsmath}
\usepackage{bbm}
\usepackage{csquotes}
\usepackage{colortbl}
\usepackage{xcolor}
\usepackage{caption}
\usepackage{multirow}
\usepackage[export]{adjustbox}
\usepackage{soul}
\definecolor{high}{HTML}{D4EDDA}  
\definecolor{low}{HTML}{F8D7DA}   

\usepackage[T1]{fontenc}

\usepackage[utf8]{inputenc}

\usepackage{microtype}

\usepackage{inconsolata}

\usepackage{graphicx}

\title{Can We Reliably Rank Model Performance across Domains\\without Labeled Data?}

\author{Veronica Rammouz$^1$, Aaron Gonzalez$^1$, Carlos Cruzportillo$^1$, Adrian Tan$^1$,\\ \textbf{Nicole Beebe$^2$, Anthony Rios$^1$} \\
  $^1$The University of Texas at San Antonio\\
  $^2$Illinois Institute of Technology\\
  \texttt{\{veronica.rammouz, anthony.rios\}@utsa.edu} \\}

\begin{document}
\maketitle
\begin{abstract}
Estimating model performance without labels is an important goal for understanding how NLP models generalize. While prior work has proposed measures based on dataset similarity or predicted correctness, it remains unclear when these estimates produce reliable performance rankings across domains. In this paper, we analyze the factors that affect ranking reliability using a two-step evaluation setup with four base classifiers and several large language models as error predictors. Experiments on the GeoOLID and Amazon Reviews datasets, spanning 15 domains, show that large language model–based error predictors produce stronger and more consistent rank correlations with true accuracy than drift-based or zero-shot baselines. Our analysis reveals two key findings: ranking is more reliable when performance differences across domains are larger, and when the error model's predictions align with the base model’s true failure patterns. These results clarify when performance estimation methods can be trusted and provide guidance for their use in cross-domain model evaluation.
\end{abstract}


\section{Introduction}

Evaluating NLP models across every possible domain and setting is expensive, yet performance can vary widely with changes in topic, geography, or language use. For instance, a sentiment classifier that performs well on urban reviews may degrade sharply on rural dialects; a toxicity detector trained on social media English may underperform on regional variants. Because collecting labeled data in each new domain is costly, practitioners often rely on proxy signals, such as dataset similarity or linguistic drift, to estimate whether a model will generalize\citep{chang2023characterizing,ramesh-kashyap-etal-2021-domain, pudasaini2025benchmarking, elsahar-galle-2019-annotate, sun2025adversarial, ramesh-kashyap-etal-2021-domain}. However, while these estimates may correlate with accuracy in aggregate, it remains unclear whether they can reliably indicate \emph{which} models or datasets will perform better than others in practice.

\begin{figure}
    \centering
    \includegraphics[width=\linewidth]{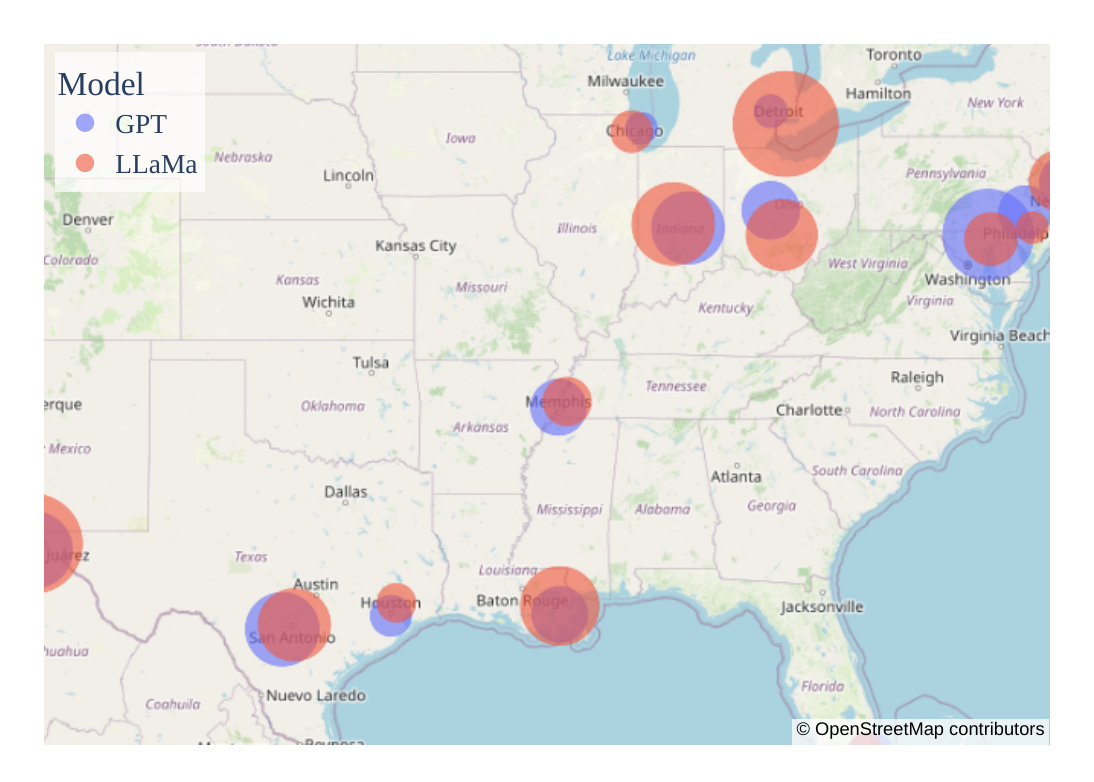}
    \caption{Geographic illustration of our estimated performance over different states in the United States of America.}
    \label{fig:map}
\end{figure}

Prior work has explored diverse strategies for predicting model performance without direct evaluation. \citet{chang2023characterizing} examined how vocabulary and structural drift affect robustness; \citet{xia2020predicting} proposed similarity metrics based on $n$-gram overlap; and transfer learning studies have analyzed how pre-trained language models such as BERT and GPT adapt to new domains through fine-tuning \citep{patankar2022train}. More recent work extends these ideas to multilingual and low-resource settings \citep{khiu2024predicting}, and to correctness estimation itself, where \citet{xiao2025generalized} showed that even large models lack reliable self-knowledge of their own correctness, motivating cross-model approaches that learn from historical prediction patterns. Together, these efforts have advanced our understanding of model generalization, yet they largely focus on predicting absolute accuracy scores. In real deployments, however, the more actionable question is often ordinal: can we rank which datasets or domains are likely to yield stronger or weaker performance, before collecting new labels? It is important to rank reliably for model selection and development purposes.

In this paper, we investigate performance prediction from a different perspective. Rather than introducing a new predictive model, we ask a more fundamental question: \emph{Can current approaches rank model performance correctly across diverse datasets and architectures?} We conduct a systematic analysis of this problem using a simple two-step framework: a base classifier (RoBERTa \cite{liu2019roberta}, a linear model, or large language models such as GPT-4o-mini \cite{openai2023gpt4} and LLaMa-3.1-8B) \cite{touvron2023llama} trained on the source task, and an auxiliary error model that estimates instance-level failures from the different base classifiers. This setup serves as a controlled environment for testing ranking reliability under varying conditions of domain shift, dataset size, and model type. By decoupling the method from the analysis, we can isolate where the predictions succeed and where they fail, providing insights into the advantages and limitations of each model.  This approach is illustrated in Figure~\ref{fig:map}: if an LLM classifier were to be applied to different cities within the United States, we would like a method that can accurately rank the performance of the for each city, with the ultimate goal of predicting where the classifier would yield poor results.

Our results reveal that existing predictors and drift-based heuristics are often unstable when ranking datasets, even when they correlate well with accuracy. Performance rankings fluctuate across architectures and domains, indicating that current estimation techniques are less robust than previously assumed. We identify key factors, such as calibration quality, representational alignment, and domain heterogeneity, that systematically drive these failures. These findings suggest that there remains a gap in knowledge for reliable ranking based on simplified grounded evaluation protocols. The present study addresses this gap in knowledge through the following:

\begin{itemize}
    \item We present the first large-scale study analyzing when and why model performance predictors succeed or fail at ranking datasets across domains.  
    \item We introduce a controlled two-step framework that enables consistent comparison of ranking reliability across model families and datasets.  
    \item We provide empirical evidence that commonly used error-prediction and drift-based methods are fragile under real distributional variation, offering new insights for improving model selection, fairness assessment, and deployment-time evaluation.
\end{itemize}

\section{Related Work}

\paragraph{Instance-Level Complexity.} Existing work examines the complexity of classification tasks through the lens of dataset complexity, or instance-level complexity measures, to understand the hardness of a given prediction task \cite{cook2025no,lorena2024trusting}. Prior work leverages a plethora of dataset-level complexity metrics (e.g., feature importance or class overlap) or model-based evaluations, (e.g., cross-validation or training loss) to allow for the prioritization of classification tasks and benchmarking. However, existing methods fail to address a fine-grained evaluation of instance hardness, specifically at the point of an NLP model's failure in a prediction task over out-of-domain datasets. We examine instance complexity through a novel two-step framework tailored to instance failures in such contexts. We employ our approach at scale to demonstrate its ranking capability across different model variants and out-of-domain datasets \cite{eberlein2025effect, zhou2023evaluating,kwon2024measuring,mallick2022matchmaker}.

\paragraph{Dataset Drift.} Much of the existing work in this area focuses on quantifying dataset drift, or covariate drift, through various metrics, such as token frequency divergences, TF-IDF vectors, and embedding-based distances \citep{ramesh-kashyap-etal-2021-domain, dredze-etal-2010-kansas, elsahar-galle-2019-annotate, ruder-etal-2017-data}. These methods have been used to predict performance degradations when moving from in-domain to out-of-domain data \citep{feldhans-etal-2021-drift}. For instance, \citet{ramesh-kashyap-etal-2021-domain} highlight the effectiveness of Jensen-Shannon divergence \cite{lin2002divergence} in assessing dataset shift, particularly in domain adaptation scenarios.


\noindent Traditional approaches to dataset drift have typically focused on holistic measures that treat the problem as a single monolithic concept, where the performance degradation is predicted based on overall divergence metrics. While useful, this approach has limitations, as it does not account for the fact that different dimensions of linguistic variation, such as changes in vocabulary, syntax, or semantics, can affect model performance in distinct ways. For example, a model's performance might degrade drastically with a change in vocabulary while being less sensitive to syntactic variations \citep{chang2023characterizing}. Prior work has shown that NLP models can exhibit high sensitivity to even small shifts in word distributions across domains, such as in the case of domain-specific jargon or regionally varied dialects \citep{yamshchikov-etal-2021-style, ramesh-kashyap-etal-2021-domain}.

Recent studies have started decomposing dataset drift into specific linguistic components, aiming to capture finer-grained distinctions. \citet{chang2023characterizing} introduce the idea of breaking down drift into vocabulary, structural, and semantic dimensions. This decomposition offers a more interpretable view of how different types of changes in the data affect model predictions. For instance, vocabulary drift focuses on divergences in content word usage, structural drift captures syntactic differences, and semantic drift identifies shifts in meaning that may not be detectable by simply looking at word distributions. This framework allows for more precise predictions of how models will generalize to new data, especially in low-resource or highly variable domains.
We leverage drift metrics as a baseline and demonstrate that they are inconsistent for ranking out-of-domain datasets.

\paragraph{Model-Based Methods.} In addition to dataset-level drift metrics, some researchers have explored the use of model-based methods, such as embedding distances, to predict how models perform on new data. Embedding-based methods, particularly those using contextualized representations from models like BERT and RoBERTa, have shown promise in capturing both syntactic and semantic shifts between datasets \citep{feldhans-etal-2021-drift}. \citet{elango-etal-2022-detect} found that fine-tuned embedding distances are particularly effective for ranking examples by expected performance, as they capture nuances that are missed by simpler, frequency-based metrics. However, while these distances can rank examples well, they are often less reliable at providing accurate performance predictions at the dataset level, particularly in out-of-domain scenarios \citep{nigenda-etal-2022-amazon}.

Finally, model-based methods are limited by focusing either on dataset-level drift or assuming access to labeled data across multiple domains. \citet{ramesh-kashyap-etal-2021-domain} and \citet{patankar2022train} highlight the need for more robust metrics that can work in low-data settings, where labeled data may only be available for a single domain. In such settings, the ability to predict model performance at the example level becomes crucial, especially when models need to be deployed in real-time applications. Our work contributes to this gap by leveraging error prediction models trained on one domain and applying them to estimate performance across multiple, unseen datasets. By predicting errors made by the base model, we introduce a second layer of predictions, which allows for more accurate estimates of model performance across a range of held-out datasets.

\paragraph{LLM-as-a-Judge.}
LLMs are increasingly employed as evaluators, so-called \emph{LLMs-as-Judges}, to replace or augment human annotation in benchmarking and meta-evaluation tasks~\cite{gu2024survey}. Early studies showed that models such as GPT-4 can approximate human preferences and provide consistent rankings across diverse tasks \citep{openai2023gpt4, nips_llm_as_a_judge}. This paradigm has since evolved into both general-purpose evaluators and fine-tuned judge models. General LLMs are directly prompted for evaluation, as in AlpacaEval \citep{alpaca_eval}, while specialized systems such as PandaLM \citep{wang2023pandalm}, JudgeLM \citep{zhu2023judgelm}, and Prometheus \citep{prometheus} fine-tune open-source backbones (e.g., Vicuna~\citep{vicuna2023}) on curated evaluation datasets to increase stability and interpretability.

Recent work extends these efforts beyond pairwise comparison toward calibrated, correctness-aware judging. Self-evaluation frameworks like Constitutional AI \citep{bai2022constitutionalaiharmlessnessai} and reasoning-centered evaluation \citep{hao2023reasoning} integrate judgment into the model’s own feedback loop, allowing iterative refinement of reasoning and decision-making. Meanwhile, meta-evaluation research highlights the need to examine how such judges generalize and whether their evaluation signals align with factual correctness rather than stylistic or lexical bias \citep{trung2024reft, aaaj}. Despite strong correlation with human preference scores, these systems often lack calibration or robustness across domains.

\begin{figure*}
    \centering
    \includegraphics[width=0.999\textwidth]{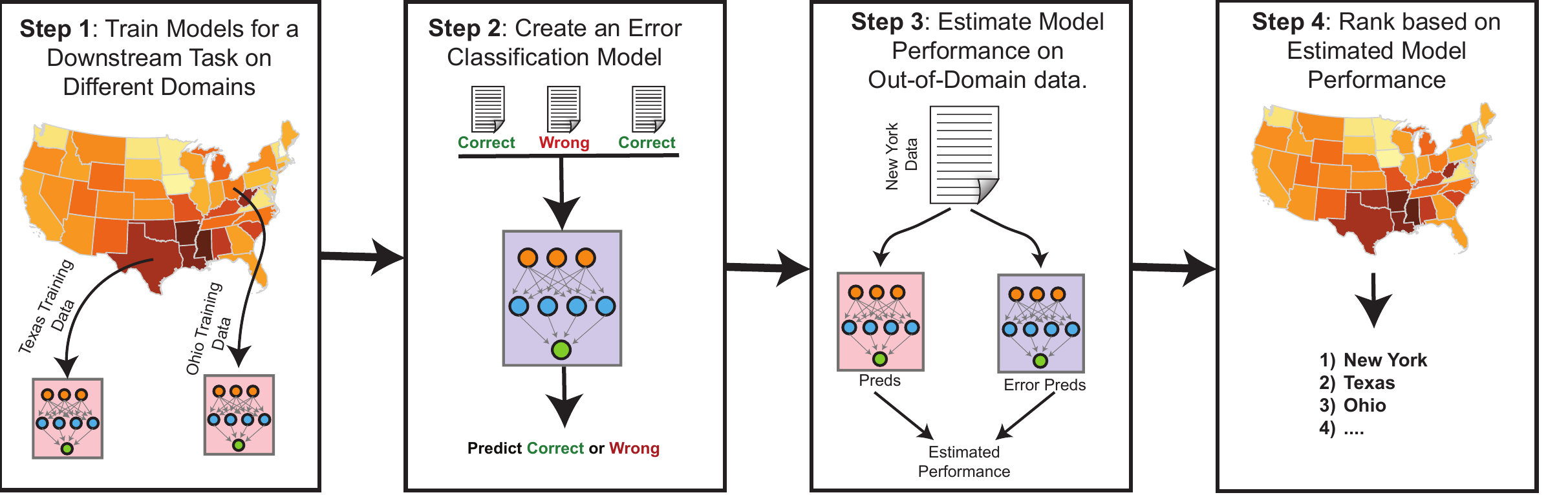}
    \caption{Overview of our two-step performance-ranking framework. In Step 1, base models are trained on source domains (e.g., different cities or product categories) for a downstream classification task. In Step 2, an error-prediction model learns to identify which instances the base model is likely to misclassify. Using these predicted errors, we estimate unlabeled accuracy on new domains and rank datasets by their expected model performance.}
    \label{fig:overview}
\end{figure*}

Most relevant to our work, recent studies on \emph{Generalized Correctness Models} argue that LLMs lack privileged “self-knowledge”: a model predicting the correctness of its own answers is typically no better than another LLM doing the same task, while training on \emph{historical} prediction data (and applying light calibration) yields cross-model correctness predictors that can match or even surpass self-emitted confidences in practice \citep{xiao2025generalized}. Rather than estimating absolute accuracy, we ask a complementary question: do such correctness estimates preserve \emph{ordinal} relationships across datasets and models? Unlike GCMs, which target calibrated probabilities, we study the robustness of \emph{ranking} under prediction error.

\section{Methodology}

Our goal is to estimate how well models will perform across unseen datasets by learning to predict model errors rather than relying on direct evaluation. As illustrated in Figure~\ref{fig:overview}, our approach proceeds in four stages: (1) train a base model on a downstream classification task, (2) collect instance-level errors from this model, (3) train a secondary model (i.e., an \emph{error predictor}) to predict those errors, and (4) use the predictions to estimate and rank the performance of models across multiple held-out datasets. We explore both supervised classifiers and LLMs in each stage, enabling us to compare traditional learning and LLM-as-a-judge paradigms under a unified framework.

Formally, let $\mathcal{D}^{(k)} = \{(x_i, y_i)\}_{i=1}^{n_k}$ denote a dataset from domain $k$, and let $f_{\theta}$ denote a model trained on labeled data $\mathcal{D}_{\text{train}}$. The model produces predictions $\hat{y}_i = f_{\theta}(x_i)$ on held-out data $\mathcal{D}_{\text{test}}$. We define the instance-level error as $e_i = \mathbb{1}[y_i \neq \hat{y}_i]$. Our objective is to estimate dataset-level performance $p^{(k)}$—e.g., accuracy or F1—without direct access to $y_i$, using the predicted probability of correctness from an auxiliary model $g_{\phi}$.

\paragraph{Step 1: Training the Base Model.}
The first step is to train or prompt a model to perform the target classification task. For smaller models, such as \texttt{RoBERTa-base}, we fine-tune parameters $\theta$ using standard cross-entropy loss:
\[
\mathcal{L}_{\text{task}}(\theta) = -\frac{1}{n} \sum_{i=1}^n y_i \log f_{\theta}(x_i).
\]
For LLMs such as \texttt{GPT-4o-mini} and \texttt{LLaMa 3.1 8B}, we instead use few-shot prompting. Each model’s outputs $\hat{y}_i$ and confidence scores (when available) serve as the foundation for error analysis in later steps. See the Appendix for the complete prompts.

\paragraph{Step 2: Learning to Predict Errors.}
Next, we train a second model $g_{\phi}$ to predict whether a given instance will be correctly classified by $f_{\theta}$. This model is conceptually similar to a \emph{generalized correctness model}~\cite{xiao2025generalized}: it estimates $g_{\phi}(x_i) \approx P(e_i = 1 \mid x_i)$ using either fine-tuned encoders or instruction-tuned LLMs prompted for binary correctness judgments (e.g., “Will the model’s answer likely be correct or incorrect?”). For smaller models, $g_{\phi}$ is trained via binary cross-entropy loss:
\begin{align*}
\mathcal{L}_{\text{error}}(\phi) = -\frac{1}{n} \sum_{i=1}^{n} [e_i \log g_{\phi}(x_i) + \\ (1 - e_i) \log (1 - g_{\phi}(x_i))].
\end{align*}
When using LLMs as error predictors, we replace supervised training with in-context exemplars of correct and incorrect predictions and prompt the LLM to judge correctness. This enables direct comparison between learned error predictors and judgment-based estimations.  See the Appendix for the complete prompts.

\paragraph{Step 3: Estimating Dataset-Level Performance.}
Given the predicted error classes, we estimate the expected accuracy on a new dataset $\mathcal{D}^{(k)}$ as
\[
\hat{p}^{(k)} = 1 - \frac{1}{|\mathcal{D}^{(k)}|} \sum_{(x_i, y_i) \in \mathcal{D}^{(k)}} \mathbbm{1}[g_{\phi}(x_i) > 0.5],
\]
which intuitively counts the number of predicted errors in the dataset. This provides a label-free proxy for model performance on unseen domains or datasets. For the LLMs, this is simply the prediction without any thresholding.

\paragraph{Performance Ranking Evaluation.} To evaluate the overall method, we measure the correlation between the predicted performance \( \hat{p}^{(k)} \) and the true performance \( p^{(k)} \) across all held-out datasets. We compute the Spearman rank \cite{spearman1961proof} correlation coefficient \( \rho \) to quantify the relationship between predicted and true performance and rank datasets,
\begin{equation*}
    \rho = \frac{\sum_{k=1}^{m} (p^{(k)} - \bar{p})(\hat{p}^{(k)} - \bar{\hat{p}})}{\sqrt{\sum_{k=1}^{m} (p^{(k)} - \bar{p})^2} \sqrt{\sum_{k=1}^{m} (\hat{p}^{(k)} - \bar{\hat{p}})^2}}
\end{equation*}
where \( \bar{p} \) and \( \bar{\hat{p}} \) are the means of the true and predicted performance scores across all datasets. High correlation values indicate that our method accurately estimates model performance across diverse datasets, even in out-of-domain scenarios. The metric we used in our experiments is Accuracy for consistency with prior work~\cite{chang2023characterizing}.

\section{Results}
We present the results of our experiments, where we evaluate the performance of models trained on different cities and test their generalization across other cities. Our key evaluation metrics are accuracy and the correlation between the estimated and true accuracy across the held-out datasets.

\paragraph{Baseline Models.} Our baseline classifiers for predicting offensive language and sentiment across both datasets include three models of increasing complexity. (1) A \textbf{Linear model}, which uses lexical and confidence-based features (details in the Appendix). (2) A \textbf{RoBERTa-base model}, fine-tuned for each task using standard cross-entropy loss (training details in the Appendix). (3) A \textbf{LLaMa-3.1-8B} model evaluated in a few-shot setting, where task-specific examples are provided through in-context prompting (the full prompt is shown in the Appendix).

\paragraph{Baselines Error Models/Methods.} We evaluate four baselines that estimate performance without labeled data, matching the names used in Tables~\ref{table1}--\ref{table2}. The \textbf{Zero-Shot Baseline} uses \texttt{GPT-4o-mini} to produce label predictions directly; we treat the model’s confidence for the predicted label as a proxy for correctness and average these probabilities within each domain to obtain an estimated accuracy (cf. \citealt{openai2023gpt4}). The \textbf{Semantic Drift} baseline measures semantic proximity between predicted and reference label texts using \textsc{SBERT} embeddings \citep{reimers2019sentence}. For each example $i$ we compute the maximum cosine similarity over candidate labels,$
s_i=\max_{y\in\mathcal{Y}}\cos\!\big(\mathbf{e}_{\hat{y}_i},\,\mathbf{e}_{y}\big),$
and estimate domain-level performance by averaging these scores,
$
\hat{A}_d=\frac{1}{N_d}\sum_{i=1}^{N_d} s_i.
$
The \textbf{RoBERTa} baseline trains a \textsc{RoBERTa}-base error model \citep{liu2019roberta} to predict whether a base classifier’s output is correct given the input and predicted label; we average predicted correctness probabilities over each domain at inference. The \textbf{Linear Model} baseline is a logistic regression error model ngram features; as with RoBERTa, we average predicted correctness to estimate accuracy. Training and optimization details for the learned error models are in the Appendix.

\paragraph{Datasets.} We use two datasets in our study: GeoOLID and Amazon Reviews 2023.

\paragraph{\textit{GeoOLID}.} The GeoOLID dataset was introduced by \citet{lwowski2022measuring} and used to investigate the geographic performance disparities of offensive language classifiers. GeoOLID contains over 14,000 annotated examples of tweets from 15 geographically and demographically diverse cities across the United States. The regional variability of language use presented in this dataset allows for studying the effect of linguistic and topical differences on model accuracies. We leverage these variations to test whether our base classifier models can accurately predict offensive language when trained on one city.

\begin{figure}[t]
    \centering
    \includegraphics[width=0.48\textwidth,left]{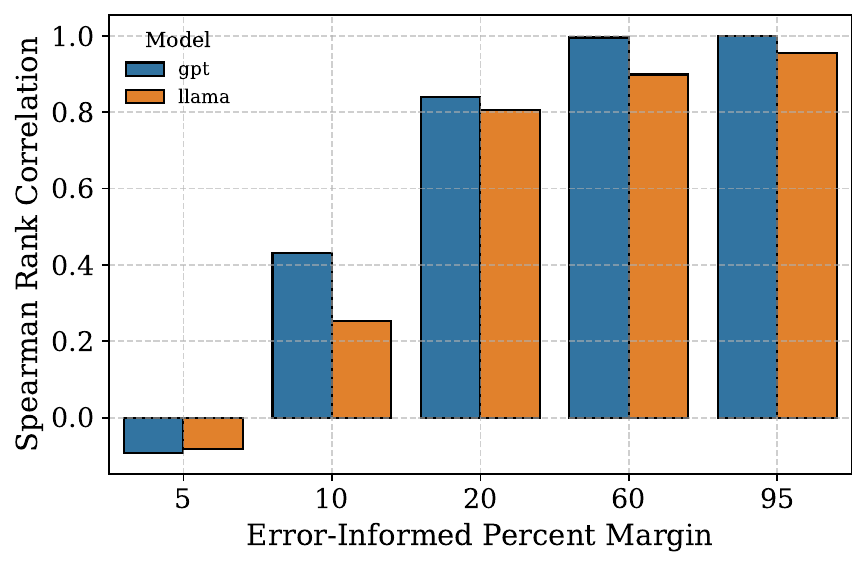}
    \caption{Figure showing the performance as a result of error-informed variations in true accuracy with a decreasing margin over datasets for the Software training category.}
    \label{fig:variations2}
\end{figure}

\begin{figure}[t]
    \centering
    \includegraphics[width=0.48\textwidth,left]{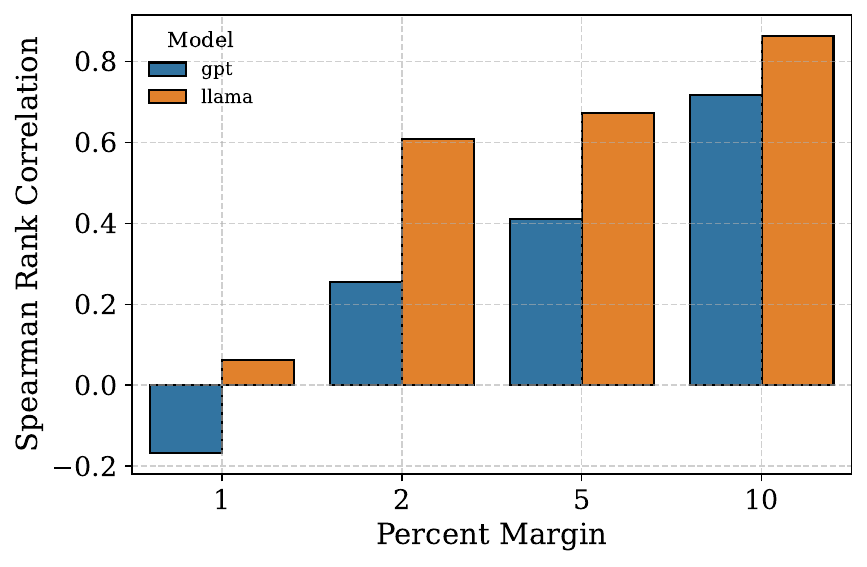}
    \caption{Figure showing the performance as a result of variations in true accuracy with a decreasing margin over datasets for the Software training category.}\vspace{-1em}
    \label{fig:variations}
\end{figure}

\begin{figure}[t]
    \centering
    \includegraphics[width=0.48\textwidth,left]{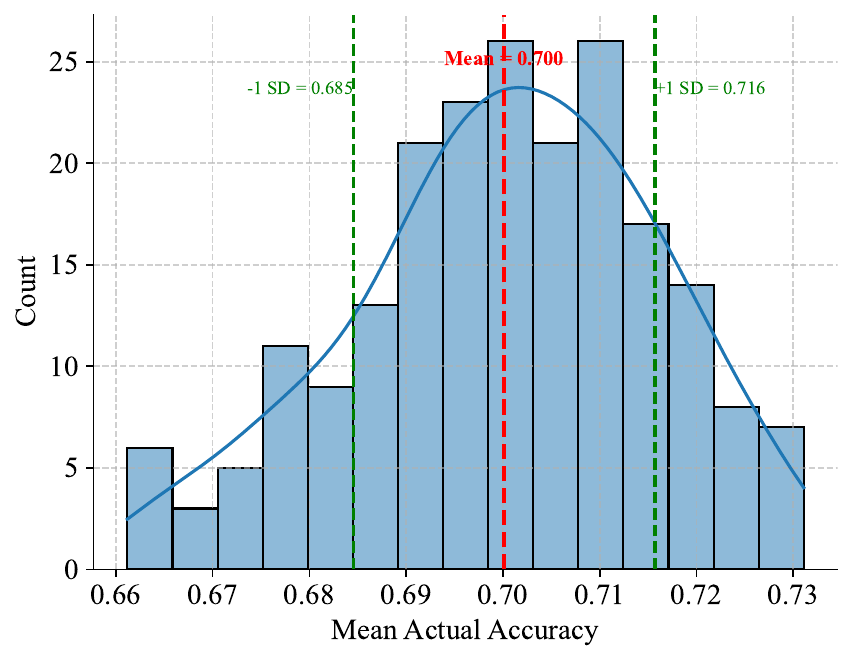}
    \caption{Histogram showing the base model accuracy ranges with their respective statistics on the Amazon Dataset.}
    \label{fig:amazon_hist}
\end{figure}

\paragraph{\textit{Amazon Reviews 2023}.} The second dataset used in our study is the Amazon Reviews 2023 dataset introduced by Hou et al. (2024). This dataset comprises over 570 million reviews and 48 million items across 33 product categories. To create a balanced and more manageable subset for analysis, we selected 15 distinct product categories, assigned a label of either "positive", "neutral", or "negative" based on the review's numerical rating, and randomly sampled 1,000 reviews for each label. Each review in the dataset is paired with detailed product metadata, including titles, features, and descriptions. This sampled subset facilitates exploring the relationship between textual reviews and item metadata for recommendation and retrieval tasks. We can use it to understand how sentiment (product rating) prediction models perform across different products.

\paragraph{Variation Study}
Figures~\ref{fig:variations2} and~\ref{fig:variations} present variation studies that analyze how ranking reliability changes when we systematically (synthetically at random) increase the difference in true accuracy between the best and worst performing domains.  These figures are on the Amazon dataset; see the Appendix for GeoOLID results that follow a similar pattern. Intuitively, we start with completely correct predictions for one city/domain, and then for each subsequent city/domain, we add errors to the predictions. We explore two conditions: (1) errors are injected at random across all instances, and (2) errors are added only to examples where the error model originally predicted a mistake. These two settings allow us to test whether ranking correlations remain stable when the true accuracy gap of what we are trying to estimate widens uniformly.

\begin{figure}[t]
    \centering
    \includegraphics[width=0.48\textwidth,left]{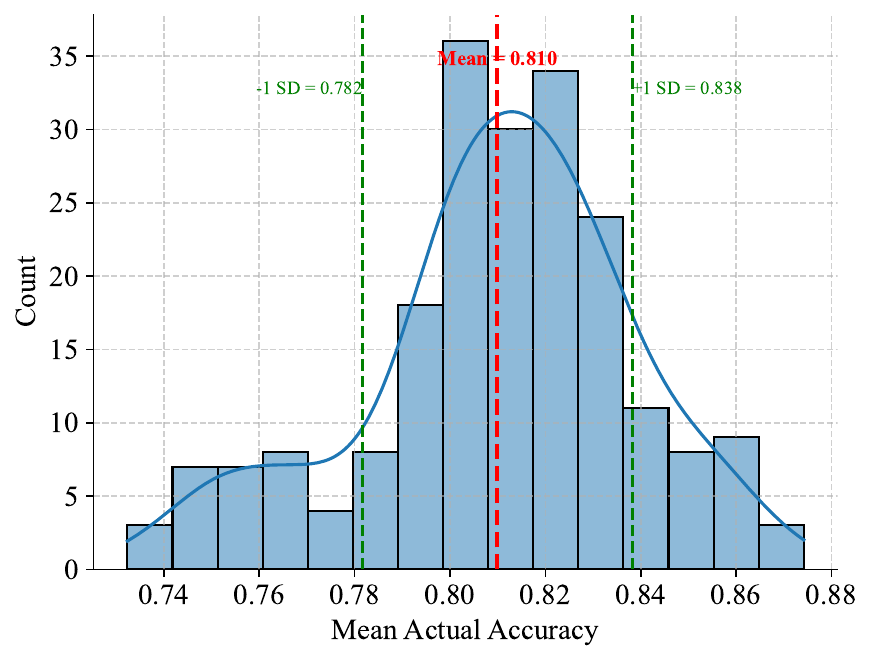}
    \caption{Histogram showing the base model accuracy ranges with their respective statistics on the GeoOLID Dataset.}\vspace{-1em}
    \label{fig:geoOLID_hist}
\end{figure}

\begin{table*}[ht]
\centering
\resizebox{\textwidth}{!}{%
\begin{tabular}{lrrrrrrrrrrrrrrrr}
\toprule
 & Bal. & Chi. & Col. & Det. & EP & Hou. & Ind. & LA & Mem. & Mia. & NO & NY & Phi. & Pho. & SA & \textbf{AVG} \\
\midrule
\multicolumn{17}{l}{\textbf{Baselines}} \\
\midrule
Semantic Drift & -.004 & .420 & -.145 & -.170 & -.001 & .096 & -.114 & .177 & -.283 & .062 & .300 & .312 & .097 & .017 & -.129 & \textbf{.042} \\ 
Zero-Shot Baseline & .256 & .254 & .277 & .234 & .254 & .273 & .289 & .281 & .262 & .231 & .234 & .205 & .252 & .242 & .253 & \textbf{.253} \\ 
\midrule
\multicolumn{17}{l}{\textbf{Performance Estimation Approach}} \\
\midrule
RoBERTa & .472 & .512 & .389 & .301 & .386 & .471 & .449 & .516 & .270 & .512 & .411 & .316 & .576 & .289 & .412 & \textbf{.419} \\
Linear Model & \cellcolor{low}-.029 & -.038 & .077 & .286 & .072 & .030 & .212 & -.018 & .123 & .251 & .185 & .169 & .132 & .136 & .301 & \textbf{.126} \\ 
\midrule
GPT-4o-mini & .599 & .556 & .494 & .525 & .567 & .537 & .539 & .611 & .448 & .549 & .556 & .575 & .490 & .462 & .401 & \textbf{.527} \\
LLaMa-3.1-8B & .583 & .580 & .640 & .662 & .535 & .690 & .687 & .657 & .611 & .609 & .627 & .631 & .658 & .577 & .620 & \textbf{.625} \\
Gemma-3-12B-it & .619 & .623 & .654 & .682 & .586 & \cellcolor{high}.707 & .689 & .683 & .562 & .608 & .625 & .662 & .656 & .580 & .602 & \textbf{.635} \\
Qwen-3-32B & .285 & .369 & .371 & .441 & .314 & .316 & .388 & .273 & .300 & .436 & .142 & .216 & .218 & .307 & .339 & \textbf{.314} \\
\bottomrule
\end{tabular}}
\caption{Spearman correlation between estimated performance and true performance when training on each of the different cities, using the following acronyms for clarity: \textbf{Bal.}: Baltimore, \textbf{Chi.}: Chicago, \textbf{Col.}: Colorado, \textbf{Det.}: Detroit, \textbf{EP}: El Paso, \textbf{Hou.}: Houston, \textbf{Ind.}: Indianapolis, \textbf{LA}: Los Angeles, \textbf{Mem.}: Memphis, \textbf{Mia.}: Miami, \textbf{NO}: New Orleans, \textbf{NY}: New York, \textbf{Phi.}: Philadelphia, \textbf{Pho.}: Phoenix, \textbf{SA}: San Antonio.}
\label{table1}
\end{table*}

\begin{table*}[ht]
\centering
\resizebox{\textwidth}{!}{%
\begin{tabular}{lrrrrrrrrrrrrrrrr}
\toprule
 & AB & AP & BP & DM & GC & GGF & HP & HPC & MT & MI & OP & PS & SW & SB & VG & \textbf{AVG} \\
\midrule
\multicolumn{17}{l}{\textbf{Baselines}} \\
\midrule
Semantic Drift & .451 & .641 & .348 & .325 & -.051 & .176 & .405 & .504 & -.164 & .303 & .571 & .582 & .014 & .035 & .390 & \textbf{.302} \\ 
Zero-Shot Baseline & .162 & .145 & .185 & .211 & .099 & .195 & .149 & .174 & .197 & .154 & .140 & .147 & .118 & .152 & .140 & \textbf{.157} \\ 
\midrule
\multicolumn{17}{l}{\textbf{Performance Estimation Approach}} \\
\midrule
RoBERTa & -.084 & -.270 & .012 & .141 & -.021 & -.175 & .051 & -.013 & .133 & .112 & -.125 & -.35 & -.172 & .032 & -.125 & \textbf{-.057} \\
Linear Model & \cellcolor{high}.588 & .436 & .468 & .209 & .166 & .374 & .207 & .527 & .086 & .392 & .475 & .494 & .200 & .208 & .354 & \textbf{.346} \\ 
\midrule
GPT-4o-mini & -.086 & -.079 & -.030 & -.123 & .174 & -.048 & -.056 & -.072 & -.193 & -.143 & -.131 & -.141 & -.236 & -.076 & -.045 & \textbf{-.086} \\
LLaMa-3.1-8B & .353 & .467 & .477 & .430 & \cellcolor{high}.588 & .574 & .275 & .397 & .253 & .395 & .476 & .535 & .005 & .321 & .331 & \textbf{.392} \\
Gemma-3-12B-it & -.119 & .145 & .044 & -.119 & .054 & .039 & -.049 & .010 & -.145 & .030 & .039 & .018 & -.242 & -.050 & .108 & \textbf{-.015} \\
Qwen-3-32B & .124 & \cellcolor{low}-.347 & -.283 & -.112 & -.308 & -.130 & -.301 & -.111 & -.112 & -.297 & -.022 & -.229 & -.035 & -.135 & -.027 & \textbf{-.139} \\
\bottomrule
\end{tabular}}
\caption{Spearman correlation between estimated performance and true performance when training on each of the different categories, using the following acronyms for clarity: \textbf{AB}: All Beauty, \textbf{AP}: Appliances, \textbf{BP}: Baby Products, \textbf{DM}: Digital Music, \textbf{GC}: Gift Cards, \textbf{GGF}: Grocery and Gourmet Food, \textbf{HP}: Handmade Products, \textbf{HPC}: Health and Personal Care, \textbf{MT}: Movies and TV, \textbf{MI}: Musical Instruments, \textbf{OP}: Office Products, \textbf{PS}: Pet Supplies, \textbf{SW}: Software, \textbf{SB}: Subscription Boxes, \textbf{VG}: Video Games.}
\vspace{-1em}
\label{table2}
\end{table*}

In the random setting (Figure~\ref{fig:variations}, correlations rise gradually as the difference in true accuracy increases, showing that larger performance gaps make it easier to identify correct rankings. When errors are injected based on the error model's original mistakes (Figure~\ref{fig:variations2}), the ranking correlations remain more stable, suggesting that the model’s internal error patterns capture some real structure in where performance decays. Together, these results indicate that reliable ranking depends not only on the scale of accuracy differences but also on how those differences relate to model-specific errors.

Figures~\ref{fig:geoOLID_hist} and~\ref{fig:amazon_hist} display histograms of the true performance distributions for the Amazon and GeoOLID datasets, respectively. Each histogram shows the mean and one standard deviation range of base model accuracies across all domains. The Amazon dataset (mean $=0.70$, SD $=0.015$) has a narrow and concentrated accuracy range, while GeoOLID (mean $=0.81$, SD $=0.028$) exhibits a wider and more uniform distribution. These differences explain why ranking results are stronger for GeoOLID: when performance values are more spread out, correlations between predicted and true rankings are easier to recover. In contrast, the Amazon accuracies are tightly clustered, so even small estimation errors can change the rank order.

\paragraph{Real Data Results.} Tables~\ref{table1} and~\ref{table2} present the Spearman rank correlation between estimated and true model performance across different domains. Each value represents how well the ranking produced by the error-prediction method aligns with the actual ranking of accuracies across cities in GeoOLID or product categories in Amazon Reviews. The average scores at the rightmost column are computed across all training domains. For comparison, the baseline results are the average of three approaches: semantic drift, covariate shift, and a zero-shot baseline. These baselines capture how similarity-based or unsupervised methods perform when no explicit error modeling is used.

Table~\ref{table1} reports results on GeoOLID. The baseline measures yield weak or inconsistent results, with average correlations below .30 across all three baselines. In contrast, the error-prediction approaches show much stronger and more stable correlations. The large language model predictors perform best, with LLaMa-3.1-8B and Gemma-3-12B-it achieving average correlations of .625 and .635, respectively. These models capture consistent ranking differences across cities such as Houston (.707) and Memphis (.562), indicating that ranking reliability improves when true accuracy differences are larger and geographically structured. The smaller RoBERTa and linear models perform less consistently, averaging .419 and .126, respectively, suggesting that they capture less of the structure in regional variation.

Table~\ref{table2} shows results for the Amazon product review dataset. Here, ranking performance is notably weaker, with most correlations close to zero. The Linear model and LLaMa-3.1-8B again perform best, with average correlations of .346 and .392, while other models show low or negative values. This pattern aligns with the narrow accuracy distribution seen in Figure~\ref{fig:amazon_hist}, where accuracies cluster near 0.70 with a standard deviation of only .015. When the range of true performance is small, small prediction errors can easily change rank order, leading to unstable results. In contrast, Figure~\ref{fig:geoOLID_hist} shows a wider spread in GeoOLID accuracies (mean $=0.81$, SD $=0.028$), producing stronger signals for ranking. The relatively strong performance of the Linear model aligns with the Semantic Drift baseline, as both rely on shallow lexical or confidence-based cues that generalize well across categories. These features capture broad sentiment polarity and stylistic consistency across product types, allowing both methods to recover smooth, monotonic estimates even when absolute performance differences are minimal.

Together, these findings align with the variation studies in Figures~\ref{fig:variations} and~\ref{fig:variations2}. When we artificially increased the difference in true accuracy across domains, Spearman correlations rose from roughly .2 to .8 in GeoOLID-like conditions, especially when variations matched the model’s original errors. This supports the hypothesis that ranking reliability depends jointly on the magnitude of true performance variation and the alignment between predicted and actual error patterns. The higher and more consistent correlations in GeoOLID, compared to the tightly clustered results in Amazon, confirm that ranking becomes more reliable when underlying performance differences are broader and systematically captured by the error models. As shown in Appendix Table~\ref{tab:appendix-category-errresults}, the Linear and \texttt{GPT-4o-mini} error models achieve the highest accuracies across datasets (.734 and .825 for Amazon and GeoOLID, respectively), demonstrating that even simple models can effectively capture structured error variation. These results reinforce that performance prediction benefits from both model-informed error structure and sufficient spread in true accuracies, which together enable more stable ranking across domains.

\paragraph{Implications.} Ranking performance depends on two main factors: the size of true performance differences across domains and the accuracy of the error predictions. When these conditions hold, rankings are more stable and reliable. We also find that the accuracy of the error models correlates with overall task performance, suggesting that better error estimation supports more trustworthy model evaluation without labels.

\section{Conclusion}
In this paper, we introduced a framework for predicting model performance across diverse datasets using paired base and error models. Base classifiers are first trained on source domains, and secondary error models are trained to predict where those classifiers will fail on unseen data. We evaluated this dual-model setup across multiple datasets and architectures, measuring how well the predicted rankings align with true performance without using labeled evaluation data.

Our findings show that reliable performance ranking depends on two interacting factors: the amount of true variation in accuracy across domains and the alignment between predicted and actual error patterns. In datasets such as GeoOLID, where accuracies are more widely distributed, large language model predictors like Gemma-3-12B-it and LLaMa-3.1-8B achieved average correlations above .62, effectively recovering the relative ordering of model performance. In contrast, when accuracies are tightly clustered, as in the Amazon dataset (mean $=0.70$, SD $=0.015$), ranking reliability declines, and simpler models such as the Linear error model perform competitively by capturing lexical and confidence-based regularities similar to those exploited by the Semantic Drift baseline. These results suggest that strong ranking performance does not necessarily require complex modeling capacity but benefits most from structured variation in the underlying task space.

Taken together, our results establish a foundation for label-free performance estimation that can guide model selection and deployment across domains. By quantifying when and how model predictions can be used to rank expected accuracy, this framework provides a scalable and interpretable alternative to exhaustive retraining and evaluation. Future work will extend this analysis to generation and multimodal tasks, where the sources of variation and error alignment may differ, and investigate how task-specific priors can improve ranking stability across heterogeneous datasets.

\section*{Acknowledgements}
This material is based upon work supported by the National Science Foundation (NSF) under Grant No.~2145357. This work was also supported in part by the U.S. Army Combat Capabilities Development Command (DEVCOM), the U.S. Army Research Laboratory (ARL), ARL South at the University of Texas at San Antonio (UTSA), and the Army Educational Outreach Program(AEOP).

\section{Limitations}
This study has several limitations that outline the scope of our analysis rather than fundamental weaknesses. First, the reliability of ranking depends on how well the error models capture model failures. These predictors may not represent all sources of variation in generalization, especially when data are highly imbalanced or differ in structure. Second, our experiments focus on classification tasks and two datasets, which limits how far the findings can be extended to other types of problems. Third, some domains in our study are more uniform than others, which may influence how correlations appear. Future work can test whether these patterns hold in more diverse domains, larger model families, or tasks such as text generation and multi-modal learning.

\bibliography{custom,anthology}

\appendix

\section{Appendix}
\label{sec:appendix}

\subsection{RoBERTa Model Hyper-parameters}
We set the tokenizer for RoBERTa to \texttt{roberta-base}, learning rate of the base model to \texttt{1e-5}, learning rate of the error model to \texttt{5e-5}, max length to 128, base model epochs to 20, error model epochs to 20, training batch size to 16, evaluation batch size to 32, warmup steps to 500, weight decay to .01, train-test split seed to 42, and internal evaluation size to .1.

\subsection{Linear Model}
Based on a comparison of prediction accuracies between the Count and TF-IDF vectorizers, the selected linear classifier model used TF-IDF vectorization with stop-word elimination and unigram tokenization. The linear error model employed TF-IDF vectorization with stop-word elimination and bigram tokenization. Bigrams were selected based on having the highest Spearman rank coefficients for error prediction of other models, e.g., RoBERTa and LLMs, when compared with unigram and trigram tokenization. 

\begin{table}[t]
\centering
\small
\setlength{\tabcolsep}{8pt} 
\renewcommand{\arraystretch}{1.1} 

\resizebox{\linewidth}{!}{\begin{tabular}{lccc}
\toprule
\textbf{Base Model} & 
\textbf{Dataset} & 
\textbf{Average Acc.} \\
\midrule
 RoBERTa       & Amazon &  .730  \\
 RoBERTa       & GeoOLID & .878  \\
  Linear      & Amazon &  .574  \\
 Linear      & GeoOLID &  .790  \\
 GPT-4o-mini & Amazon  &  .751  \\
 GPT-4o-mini & GeoOLID  & .782 \\
 LLaMa-3.1-8B & Amazon  & .745 \\
 LLaMa-3.1-8B & GeoOLID  & .793 \\

\bottomrule
\end{tabular}}
\caption{Base models' average accuracy.}
\label{tab:appendix-category-results2}
\vspace{0.5em}
\end{table}

\subsection{LLM Prompts}

\paragraph{(a.1) Base Model System Prompt for GeoOLID}
\begin{footnotesize}
\begin{quote}
\textbf{SYSTEM}\\
You are a helpful assistant. Your task is to provide sound judgement on the nature of the text that will be provided to you. If you think the text is offensive, please say 'offensive'. If you think the text is not offensive, please say 'not offensive'. Take into consideration human tone. Here are some examples:

\vspace{2mm}
\textbf{User/Assistant Few-shot Examples}\\
Text: \{text\} \newline
Label: \{label\}

\vspace{2mm}
\textbf{User}\\
Text: \{text\} \newline

\end{quote}
\end{footnotesize}

\paragraph{(a.2) Base Model System Prompt for Amazon}
\begin{footnotesize}
\begin{quote}
\textbf{SYSTEM}\\
You are a helpful assistant. Your task is to provide sound judgement on the nature of the text that will be provided to you. Your task is sentiment analysis. If you think the text is positive, please say 'positive'. If you think the text is neutral, please say 'neutral'. If you think it is negative, please say 'negative'. Only say 'positive', 'negative', or 'neutral'. Here are some examples:

\vspace{2mm}
\textbf{User/Assistant Few-shot Examples}\\
\\
Text: \{text\} \newline
Label: \{label\}

\vspace{2mm}
\textbf{User}\\
Text: \{text\} \newline

\end{quote}
\end{footnotesize}

\paragraph{(b) Error Model System Prompt}
\begin{footnotesize}
\begin{quote}
\textbf{SYSTEM}\\
You are a helpful assistant. Your task is to check whether our prediction is an error or not based on sound judgment about the nature of the text. You will check if the predicted label is correct for the given text. If you think the predicted label is correct, please say 'correct'. If you think the predicted label is wrong, please say 'error'. Only say 'error' or 'correct'. Please note that the order of the examples does not matter - the content matters. Here are some examples:

\vspace{2mm}
\textbf{User/Assistant Few-shot Examples}\\
Text: \{text\} -> Predicted Label: \{label\} \newline
Error Label: \{label\}

\vspace{2mm}
\textbf{User}\\
Text: \{text\} -> Predicted Label: \{label\} \newline

\end{quote}
\end{footnotesize}


\section{Additional Results}

\begin{table}[t]
\centering
\small
\setlength{\tabcolsep}{8pt} 
\renewcommand{\arraystretch}{1.1} 

\resizebox{\linewidth}{!}{\begin{tabular}{lccc}
\toprule
\textbf{Error Model} & \textbf{Dataset}  & \textbf{Average Acc.} & \textbf{Average F1} \\
\midrule
 Linear & Amazon  &  .734 & --\\
 Linear & GeoOLID  &  .825  & --\\
 GPT-4o-mini & Amazon  & .674 & .750 \\
 GPT-4o-mini & GeoOLID  & .737 & .812 \\
 LLaMa-3.1-8B & Amazon  & .716 & .780 \\
 LLaMa-3.1-8B & GeoOLID  & .644 & .728 \\
 Gemma      & Amazon & .632 & .453 \\
 Gemma      & GeoOLID & .687 & .487 \\
 Qwen       & Amazon & .372 & .457 \\
 Qwen       & GeoOLID & .644 & .203 \\
\bottomrule
\end{tabular}}
\caption{Error models' average accuracy and average F1 scores per dataset.}
\label{tab:appendix-category-errresults}
\vspace{0.5em}
\end{table}

Table~\ref{tab:appendix-category-results2} reports the average accuracy of the base models across the Amazon and GeoOLID datasets. Performance varies by model and dataset, with \texttt{RoBERTa} achieving the highest overall accuracies (.730 on Amazon and .878 on GeoOLID). The \texttt{Linear} model performs lowest on both datasets (.574 and .790), while the large language models (\texttt{GPT-4o-mini} and \texttt{LLaMa-3.1-8B}) fall in between, averaging around .75 on Amazon and .79 on GeoOLID. These results show that base model performance differs across datasets and model types, with GeoOLID generally yielding higher accuracies.

Table~\ref{tab:appendix-category-errresults} presents the average accuracy and F1 scores for the error prediction models. The \texttt{Linear} error model achieves the highest accuracies overall (.734 on Amazon and .825 on GeoOLID), followed by \texttt{GPT-4o-mini} (.674 and .737) and \texttt{LLaMa-3.1-8B} (.716 and .644). \texttt{Gemma} and \texttt{Qwen} perform worse across both datasets, with notably low F1 scores on GeoOLID (.487 and .203). These results indicate that while larger models can learn complex patterns, simpler error models can produce stable and competitive accuracy estimates across datasets.

Figures~\ref{fig:variations2} illustrate the relationship between performance ranking and the margin of true accuracy differences for the Detroit training category in GeoOLID. The first figure shows how Spearman rank correlation decreases as the margin between true accuracies narrows, demonstrating that ranking reliability weakens when performance gaps are smaller. The second figure applies the same analysis under an error-informed setting, where performance variation is guided by the model’s predicted error distribution. In this case, correlations remain higher across decreasing margins, suggesting that model-informed variation better preserves ranking consistency. Together, the figures highlight how performance differences and error structure influence correlation stability across datasets.

\begin{figure}[t]
    \centering
    \includegraphics[width=0.48\textwidth,left]{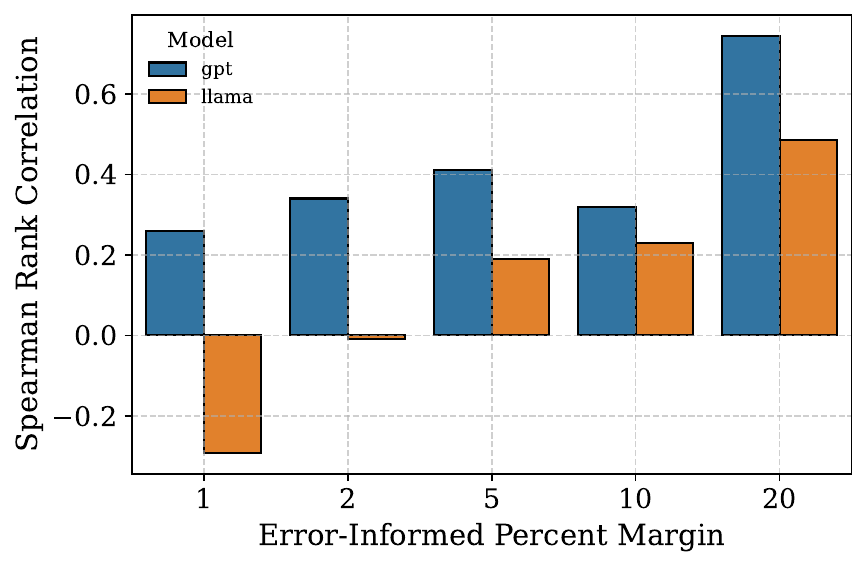}
    \caption{Figure showing the performance as a result of variations in true accuracy with a decreasing margin over datasets for the Detroit training category.}
    \label{fig:variations2}
\end{figure}

\begin{figure}[t]
    \centering
    \includegraphics[width=0.48\textwidth,left]{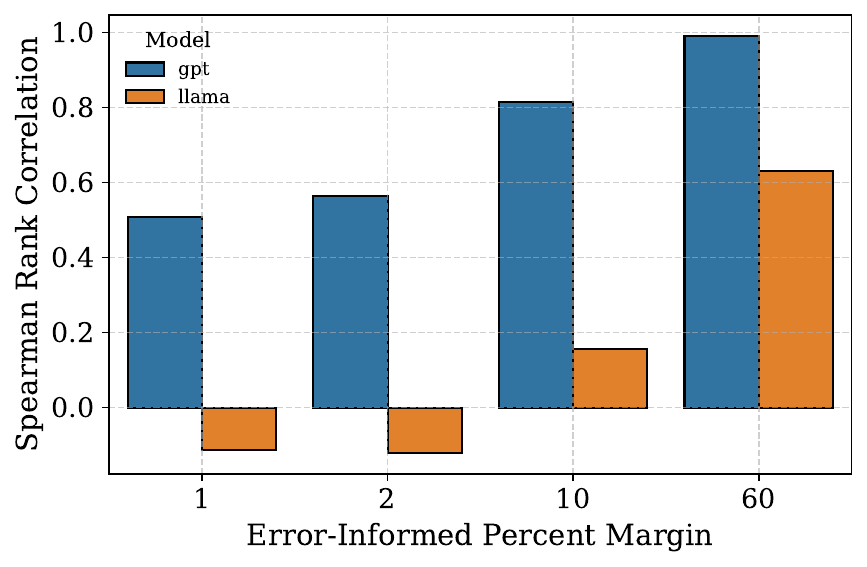}
    \caption{Figure showing the performance as a result of error-informed variations in true accuracy with a decreasing margin over datasets for the Detroit training category.}
    \label{fig:variations2}
\end{figure}

\section{Use of AI}
We used AI only to help write the paper, e.g., to fix grammar and sentences. All ideas are our own.

\end{document}